\documentclass[letterpaper,journal]{IEEEtran}
\usepackage{amsmath,amsfonts}
\usepackage{cite}
\usepackage{amsmath,amssymb,amsfonts}
\usepackage{graphicx}
\usepackage{url}

\usepackage{booktabs}
\usepackage{multirow}
\usepackage{makecell}
\usepackage{array}
\usepackage{adjustbox}
\usepackage[table]{xcolor}

\usepackage{algorithm}
\usepackage{algpseudocode}

\begin{document}

\title{SpatialFly: Implicit 3D Prior-Guided Visual Reparameterization \\for Continuous UAV Vision-and-Language Navigation}

\author{
Wen Jiang,
Kangyao Huang,
Li Wang,
Wang Xu,
Wei Fan,
Jinyuan Liu,
Shaoyu Liu,
Hanfang Liang,\\
Hongwei Duan,
Bin Xu,
Xiangyang Ji,~\IEEEmembership{Senior Member, IEEE}
and Huaping Liu,~\IEEEmembership{Fellow,~IEEE}
\thanks{
This work was supported by the National Natural Science Foundation of China under Grant No. 52502496, U22B2052 and the Natural Science Foundation of Chongqing, China under Grant No. CSTB2025NSCQ-GPX0413, and the National High Technology Research and Development Program of China under Grant No. 2020YFC1512501. (Corresponding authors: Bin Xu and Xiangyang Ji)}
\thanks{
Wen Jiang, Li Wang, Wei Fan, Hongwei Duan, and Bin Xu are with the School of Mechanical Engineering, Beijing Institute of Technology, Beijing 100081, China. Li Wang is also with the Chongqing Innovation Center, Beijing Institute of Technology, Chongqing 401120, China. (Corresponding author: Bin Xu, e-mail: bitxubin@bit.edu.cn).}
\thanks{Wen Jiang, Li Wang, Wei Fan, Hongwei Duan, and Bin Xu are with the School of Mechanical Engineering, Beijing Institute of Technology, Beijing 100081, China. (e-mail: 3120235086@bit.edu.cn, wangli\_bit@bit.edu.cn,fanweixx@bit.edu.cn, 3220250437@bit.edu.cn, bitxubin@bit.edu.cn, }
\thanks{Kangyao Huang is with the Department of Computer Science and Technology, Tsinghua University, Beijing 100084, China.(e-mail: huangky22@mails.tsinghua.edu.cn)}

\thanks{Wang Xu is with Tsinghua University, Beijing 100084, China.(e-mail:xwjim812@126.com)}

\thanks{Li Wang is also with the Chongqing Innovation Center, Beijing Institute of Technology, Chongqing 401120, China.}

\thanks{Xiangyang Ji is with the Department of Automation, Tsinghua University, Beijing 100084, China.(e-mail: xyji@tsinghua.edu.cn)}
\thanks{Jinyuan Liu is with the School of Software, Dalian University of Technology, Dalian 116024, China.(e-mail: jinyuanliu@dlut.edu.cn)}
\thanks{Shaoyu Liu is with the School of Artificial Intelligence, Xidian University, Xi'an 710071, China.(e-mail: 23171110721@stu.xidian.edu.cn)}
\thanks{Hanfang Liang is Jianghan University, Wuhan 430056, China.(e-mail: lhf.liang@gmail.com)}
\thanks{Huaping Liu is with the Department of Computer Science and Technology, Tsinghua University, Beijing 100084, China (e-mail: hpliu@tsinghua.edu.cn).}

}


\markboth{IEEE ROBOTICS AND AUTOMATION LETTERS,~Vol.~11, No.~6, June~2026}%
{Shell \MakeLowercase{\textit{et al.}}: A Sample Article Using IEEEtran.cls for IEEE Journals}


\maketitle

\begin{abstract}
UAVs play an important role in applications such as autonomous exploration, disaster response, and infrastructure inspection. However, UAV VLN in complex 3D environments remains challenging. A key difficulty is the structural representation mismatch between 2D visual perception and the 3D trajectory decision space, which limits spatial reasoning. To this end, we propose SpatialFly, a geometry-guided spatial representation framework for UAV VLN. Operating on RGB observations without explicit 3D reconstruction, SpatialFly introduces a geometry-guided 2D adaptive representation mechanism. Specifically, the geometric prior injection module injects global structural cues into 2D semantic tokens to provide scene-level geometric guidance. The geometry-aware reparameterization module then uses geometry-conditioned cross-modal attention and gated
residual fusion to adaptively reparameterize the visual tokens. Experimental results show that SpatialFly consistently outperforms state-of-the-art UAV VLN baselines across both seen and unseen environments, reducing NE by 4.03\,m and improving SR by 1.27\% over the strongest baseline on the unseen Full split. Additional trajectory-level analysis shows that SpatialFly produces trajectories with better path alignment and smoother, more stable motion.
\end{abstract}


\section{Introduction}
\IEEEPARstart{U}{nmanned} 
 aerial vehicle vision-and-language navigation (UAV VLN) requires UAVs~\cite{11513999, 10578307, 8372447} to understand natural language instructions and generate continuous flight decisions in 3D environments based on visual observations. Unlike ground robot navigation~\cite{11027320, 10776999, 11192056}, while UAVs mainly rely on multi-view 2D RGB observations to perceive the environment, they must produce trajectories that satisfy geometric constraints in 3D~\cite{zhang2026spatialnavleveragingspatialscene, zhang2026apexdecoupledmemorybasedexplorer,lin2025evo0visionlanguageactionmodelimplicit}. This setting introduces a structural representation mismatch between 2D perception and 3D trajectory decision making, making it difficult to maintain consistent spatial understanding and cross-view consistency~\cite{wewer2025spatialreasoningdenoisingmodels, zheng2025multimodalspatialreasoninglarge}. Therefore, although existing methods have achieved progress in ground-based or simplified scenarios, bridging the gap between 2D perception and 3D decision making remains a key challenge in UAV VLN.

\begin{figure}[t]
    \centering
    \includegraphics[width=1\linewidth]{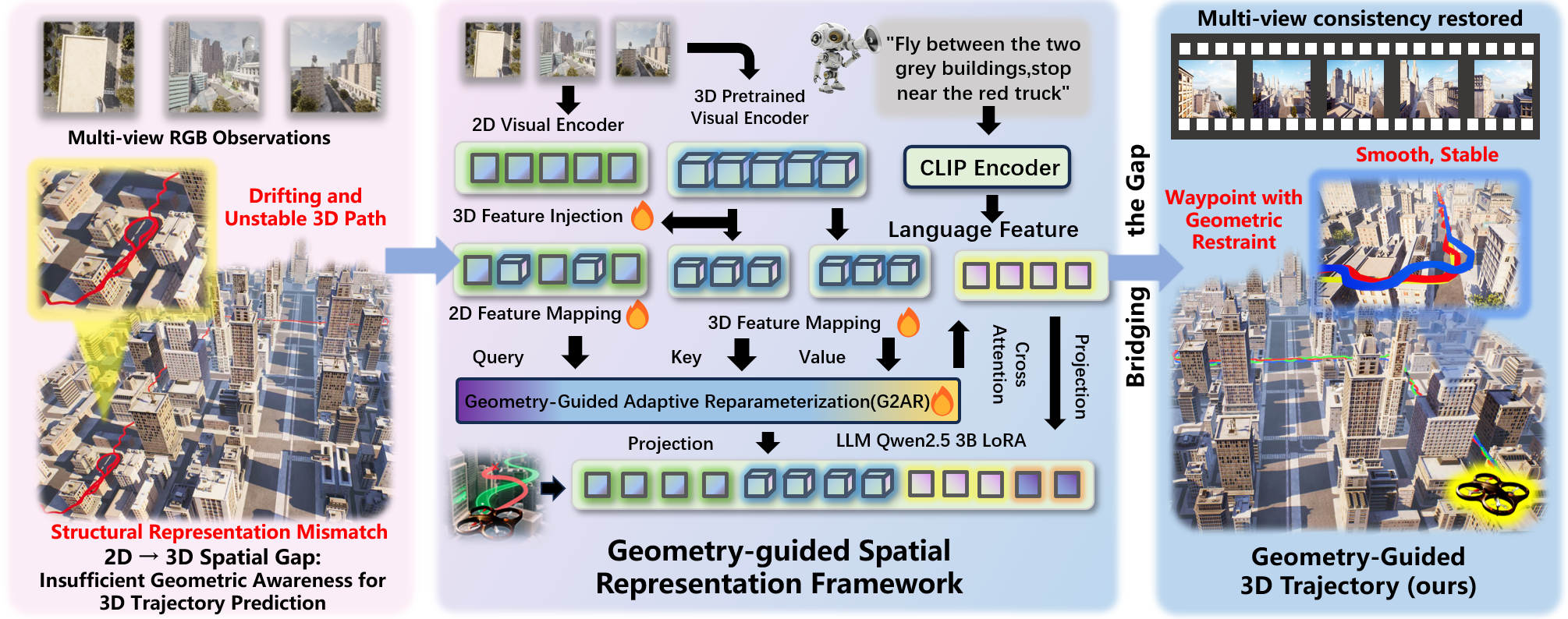}
    \caption{Motivation and overview of SpatialFly. UAV VLN suffers from a structural representation mismatch between multi-view 2D visual perception and continuous 3D trajectory decision making. Without explicit geometric cues, multi-view RGB observations often lead to inconsistent spatial understanding and unstable 3D path prediction. SpatialFly addresses this gap through geometry-guided adaptive representation mechanism, improving cross-view consistency and geometry-aware trajectory generation.}
    \label{fig:motivation}
\end{figure}

Most existing UAV VLN methods struggle to achieve stable spatial reasoning in complex 3D environments, mainly due to the lack of representations that bridge 2D visual perception and continuous 3D decision spaces. Early studies, such as AerialVLN~\cite{liu2023aerialvln} and AVDN~\cite{fan2023aerial}, simplify the continuous flight process into finite decision sequences by adopting predefined discrete action spaces~\cite{ye2026flyawareinertiaawareaerialmanipulation}. Although such discretization simplifies navigation, the reliance on fixed action sets limits generalization to unseen environments. Subsequently, some studies introduce explicit geometric modeling methods~\cite{10.1145/3757374.3771534, ZHOU2026105075}, such as bird’s-eye-view topological~\cite{s25196106} or semantic maps~\cite{canh2024objectorientedsemanticmappingreliable}, to enhance global spatial perception. With the emergence of large language models (LLMs)~\cite{YANG2026112986, cheng2024spatialrgptgroundedspatialreasoning}, semantic-level planning ability has improved, but limitations in spatial intelligence remain evident, as current vision-language models (VLMs)~\cite{chen2024spatialvlmendowingvisionlanguagemodels, zhu2025llava3dsimpleeffectivepathway, xu2026aerialvisionlanguagenavigationunified} still struggle to reliably encode 3D geometric structures. Recent studies such as TravelUAV~\cite{wang2024towards}, AutoFly~\cite{sun2026autoflyvisionlanguageactionmodeluav} and FlightGPT~\cite{cai2025flightgptgeneralizableinterpretableuav} enhance high-level semantic reasoning by leveraging large-scale vision-language models. AeroDuo~\cite{wu2025aeroduoaerialduouavbased} distributes spatial cognition through multi-UAV cooperation, while LongFly~\cite{jiang2025longflylonghorizonuavvisionandlanguage} improves navigation stability by incorporating historical trajectories and visual observations. While current methods enhance policy-level performance, they fail to resolve the structural mismatch between 2D perception and 3D decision spaces, leaving geometry-guided visual reparameterization as a critical open challenge.

To bridge this gap, we identify a key challenge in UAV VLN: the mismatch between multi-view 2D perception and continuous 3D decision making. First, existing methods lack a spatial representation that can connect 2D visual observations with 3D trajectory decisions. Second, with only 2D inputs, it is difficult to capture spatial cues such as depth, scale, and cross-view consistency. As a result, conventional 2D representations are difficult to support reliable 3D-aware reasoning for UAV navigation.

To address these challenges, we propose SpatialFly, a geometry-guided framework for UAV VLN, as shown in Fig.~\ref{fig:motivation}. Operating on RGB observations without explicit 3D reconstruction, SpatialFly introduces a geometry-guided adaptive representation mechanism. Specifically, the geometric prior injection module injects global structural cues into 2D semantic tokens to provide scene-level geometric guidance. The geometry-aware reparameterization module then reparameterizes visual tokens through geometry-conditioned cross-modal attention and gated residual fusion. Together, these designs improve the spatial consistency of visual representations and support more reliable 3D trajectory decision making in complex environments.

In summary, our main contributions are as follows:
\begin{itemize}
    \item We propose SpatialFly, a geometry-guided spatial representation framework for UAV VLN, which mitigates the structural representation mismatch between 2D visual perception and the 3D trajectory decision space.
    
    \item We introduce a geometry-guided 2D adaptive representation mechanism for RGB-only UAV VLN, which injects implicit 3D geometric priors into visual tokens and reparameterizes visual tokens through geometry-conditioned cross-modal attention and gated residual fusion.
    
    \item Experimental results show that SpatialFly consistently outperforms state-of-the-art UAV VLN baselines across both seen and unseen environments, reducing NE by 4.03\,m and improving SR by 1.27\% over the strongest baseline on the unseen Full split.

\end{itemize}

\section{Related Work}
\label{sec:related_work}
\subsection{UAV Vision-and-Language Navigation}  

UAV VLN requires UAVs to autonomously navigate in complex 6-DoF aerial spaces according to natural language instructions. Compared with ground-based VLN~\cite{yao2025navmorph, pmlr-v235-gao24p}, UAVs have greater motion freedom and larger viewpoint changes, making it harder to map visual observations to the underlying 3D space. Early studies such as AerialVLN~\cite{liu2023aerialvln} simplify the navigation process by adopting discrete action spaces. However, as the task gradually expands to large-scale outdoor environments, recent benchmarks such as OpenFly~\cite{gao2025openfly}, VLA-AN~\cite{wu2025vlaanefficientonboardvisionlanguageaction} and TravelUAV~\cite{wang2024towards} introduce large-scale trajectory data and continuous control signals, shifting the research focus toward stable trajectory prediction in continuous 3D spaces. At the methodological level, research has gradually moved from traditional cross-modal policy learning to hierarchical decision-making frameworks driven by large language models. For example, FlightGPT~\cite{cai2025flightgptgeneralizableinterpretableuav} improves the interpretability of decision reasoning by combining reinforcement learning with chain-of-thought reasoning. CityNav~\cite{lee2024citynav} and related works leverage semantic priors from large models to enhance target understanding and path planning in open environments. Besides, AeroDuo~\cite{wu2025aeroduoaerialduouavbased} coordinates multi-altitude UAVs to decompose global reasoning from local execution. LongFly~\cite{jiang2025longflylonghorizonuavvisionandlanguage} models historical observations in a temporal manner to alleviate navigation drift. However, most methods still model navigation as a direct mapping from 2D visual sequences to a 3D action space, ignoring the structural representation mismatch between visual perception and the underlying 3D decision space.

\subsection{Implicit 3D Representations for Navigation} 

Building autonomous navigation in complex 3D environments depends on creating spatial representations, which are generally categorized as explicit or implicit. Explicit methods build structured 3D maps through voxels or occupancy grids; for instance, VER\cite{liu2024volumetricenvironmentrepresentationvisionlanguage} projects multi-view visual features into a unified voxel space to improve environmental modeling, though it requires significant memory and computing power. In contrast, implicit methods model 3D structures more compactly, with works like NeRF~\cite{mildenhall2021nerf}, NICE-SLAM~\cite{zhu2024nicer}, and Co-SLAM~\cite{wang2023co} using neural implicit representations to learn scene geometry, while SPAR-7M~\cite{zhang2025flatland} shows that multi-view supervision can boost spatial reasoning from RGB input. Additionally, GeoNav~\cite{XU2026113365} highlights that UAV navigation requires multi-scale reasoning to follow complex language goals. Therefore, a major challenge remains: how to learn cross-view consistent implicit 3D representations from RGB observations and apply them to continuous 3D flight decisions.

\section{Method}
\label{sec:method}

\subsection{Overview of the SpatialFly Framework}
\label{sec:overview}

\begin{figure}
    \centering
    \includegraphics[width=1\linewidth]{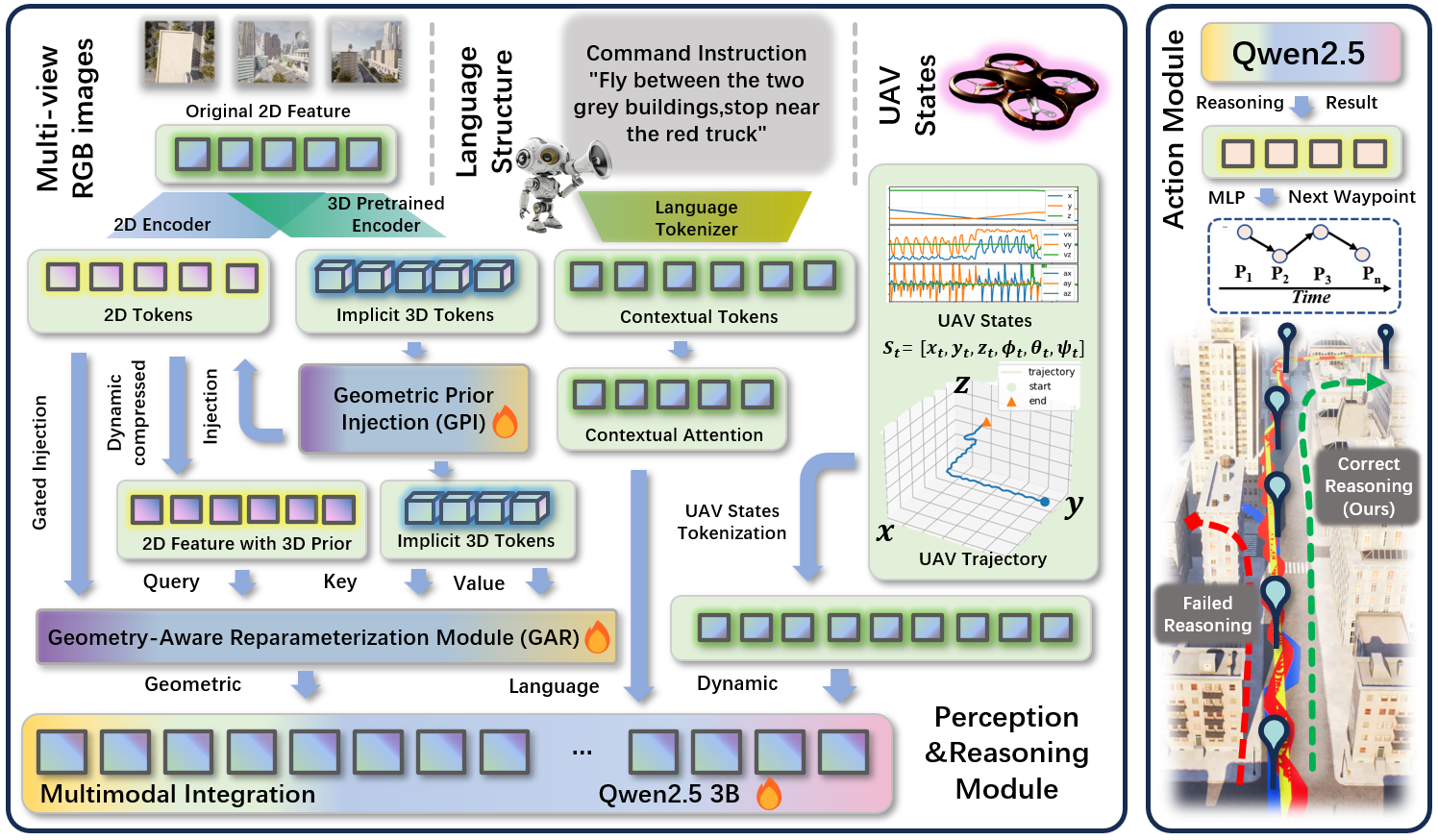}
    \caption{Overall architecture of SpatialFly. Given multi-view RGB observations, the language instruction, and the UAV state, SpatialFly extracts 2D semantic tokens and implicit 3D geometric tokens. The GPI module injects global structural cues into 2D semantic tokens for scene-level geometric guidance. The GAR module then aligns 2D semantic tokens with 3D geometric tokens through cross-modal attention and gated fusion. The aligned visual representations are then integrated with language and state tokens for downstream action prediction.}
    \label{fig:framework}
\end{figure}

As shown in Fig.~\ref{fig:framework}, we propose SpatialFly, a geometry-guided spatial representation framework for UAV VLN. Specifically, at time step \(t\), given the language instruction \(L\), the current UAV state \(S_t\), and the multi-view observation \(R_t\), SpatialFly first extracts two streams of features:
\begin{equation}
F^{2D}_t = f_{\text{2D}}(R_t), \qquad
F^{3D}_{t,\text{raw}} = f_{\text{3D}}(R_t),
\label{eq:two_stream}
\end{equation}
where \(f_{\text{2D}}(\cdot)\) is a 2D visual encoder for semantic feature extraction, and \(f_{\text{3D}}(\cdot)\) is a geometry encoder operating on multi-view RGB observations. For VGGT~\cite{wang2025vggtvisualgeometrygrounded}, we do not use its original prediction heads including pose, depth, and point cloud estimation.

Instead, we keep only its Transformer trunk and directly use the aggregated tokens from the last layer as geometric priors. These tokens provide implicit geometric cues and cross-view structural information learned during pretraining.
The geometric priors are then aligned with and injected into the 2D representations, producing geometry-enhanced fused features:
\begin{equation}
F^{\text{fuse}}_t = \mathcal{A}\!\left(F^{2D}_t,\, F^{3D}_{t,\text{raw}}\right),
\label{eq:align_module}
\end{equation}
where \(F^{\text{fuse}}_t\) keeps the same token form as \(F^{2D}_t\), so it can be directly fed into the downstream navigation head without changing its interface. \(\mathcal{A}(\cdot)\) denotes the proposed G2RA mechanism, which consists of a geometric prior injection stage and a geometry-aware reparameterization stage.
During decision making, the downstream module combines \(L\), \(F^{\text{fuse}}_t\), and \(S_t\) into a prompt, and uses a large language model (Qwen2.5 3B) to obtain a hidden representation and regress the waypoint increment:
\begin{equation}
h_t = \mathrm{LLM}\big(\mathrm{Prompt}(L, F^{\text{fuse}}_t, S_t)\big), \qquad
\Delta W_t = \mathrm{MLP}(h_t).
\label{eq:llm_head}
\end{equation}
where the instruction tokens, projected visual tokens, and state embeddings are concatenated into a unified multimodal sequence, and the resulting hidden representation \(h_t\) is used for waypoint regression. The predicted waypoint increment is then used to update the next waypoint.

\subsection{Geometry-Guided 2D Adaptive Representation Mechanism}
\label{sec:fusion}

We design a geometry-guided 2D adaptive representation mechanism (G2RA) to reduce the mismatch between 2D semantic perception and 3D trajectory decision-making. Specifically, the geometric prior injection module injects global structural cues into 2D semantic tokens to provide scene-level geometric guidance. The geometry-aware reparameterization module then aligns 2D semantic tokens with 3D geometric tokens through cross-modal attention, followed by gated residual fusion to preserve semantic discrimination.

\subsubsection{Geometric Prior Injection Module}

\label{sec:gpi}

\begin{figure}[t]
    \centering
    \includegraphics[width=0.55\linewidth]{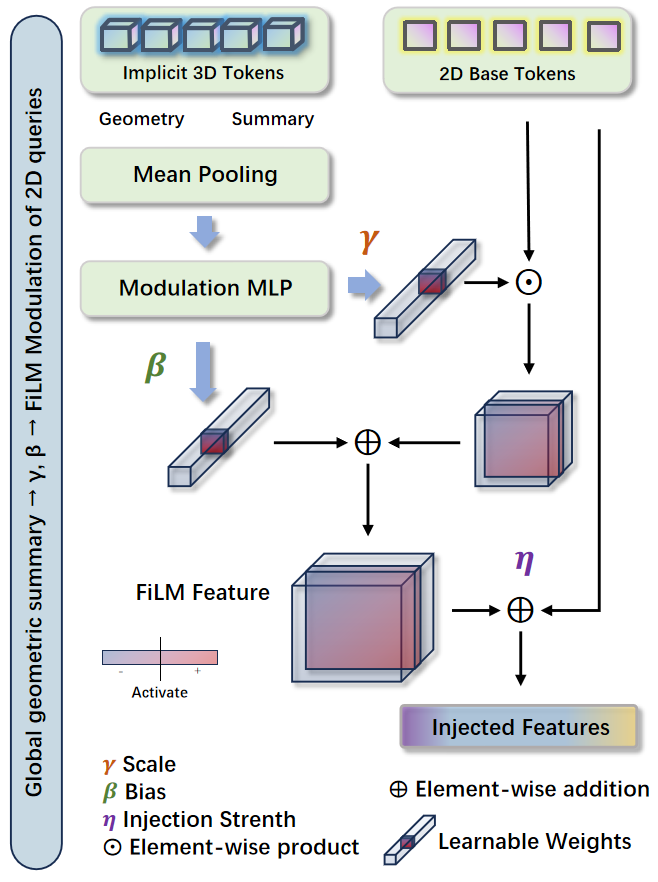}
    \caption{Illustration of the Geometric Prior Injection (GPI) module. Implicit 3D geometric tokens are first summarized by mean pooling to obtain a global geometric representation, which is then passed through a modulation MLP to generate the FiLM parameters \(\gamma\) and \(\beta\). These modulation terms are applied to the 2D base tokens in a FiLM-like manner, and the resulting features are further combined with the original tokens through a learnable injection strength \(\eta\) to produce geometry-injected representations.}
    \label{fig:model1}
\end{figure}

The geometric prior injection module (GPI) aims to inject implicit 3D geometric priors into 2D semantic tokens in a lightweight manner, so that the tokens carry global structural cues as shown in Fig.~\ref{fig:model1}. Given the 2D tokens extracted by CLIP,
$F^{2D}_t\in\mathbb{R}^{N_{2D}\times d_{\text{clip}}}$, we first obtain the base query representations via a linear projection:
\begin{equation}
Q^{\text{base}}_t = F^{2D}_t W_Q^{\text{base}} \in \mathbb{R}^{N_{2D}\times d},
\label{eq:q_base}
\end{equation}
where $W_Q^{\text{base}}\in\mathbb{R}^{d_{\text{clip}}\times d}$.

Meanwhile, VGGT~\cite{wang2025vggtvisualgeometrygrounded} (NoHead) outputs raw geometric tokens
\begin{equation}
F^{3D}_{t,\mathrm{raw}} \in \mathbb{R}^{N_{3D}\times D_{\mathrm{agg}}},
\end{equation}

To match the 2D representation space, we project them into the shared dimension $d$ and form the value sequence:

\begin{equation}
V_t = F^{3D}_{t,\mathrm{raw}} W_V + b_V \in \mathbb{R}^{N_{3D}\times d},
\label{eq:v_proj}
\end{equation}
where $W_V\in\mathbb{R}^{D_{\mathrm{agg}}\times d}$ and $b_V\in\mathbb{R}^{d}$ are learnable parameters.

\paragraph{Geometric summary and FiLM modulation}
We apply mean pooling over the token dimension of $V_t$ to obtain a global geometric summary:
\begin{equation}
g_t = \mathcal{P}(V_t)=\frac{1}{N_{3D}}\sum_{j=1}^{N_{3D}} v_{t,j}
\in\mathbb{R}^{d},
\label{eq:g_pool}
\end{equation}
and use a modulation generator $\Psi(\cdot)$ to produce affine modulation parameters:
\begin{equation}
[\gamma_t,\beta_t] = \Psi(g_t;\theta_{\Psi}), \qquad 
\gamma_t,\beta_t\in\mathbb{R}^{d},
\label{eq:film_gen}
\end{equation}
where $\Psi(\cdot)$ is a two-layer MLP. We apply a Sigmoid function to the scaling vector for stability:
$\tilde{\gamma}_t=\sigma(\gamma_t)$.

Finally, we inject the geometric priors into the 2D query tokens in a FiLM-like manner:
\begin{equation}
Q^{\text{inj}}_t
=
Q^{\text{base}}_t
+
\eta\left(
\tilde{\gamma}_t\odot Q^{\text{base}}_t
+
\beta_t
\right),
\label{eq:gpi}
\end{equation}
where $\eta$ is a learnable injection strength, $\odot$ denotes element-wise multiplication, and $\tilde{\gamma}_t,\beta_t$ are broadcast to all $N_{2D}$ 2D tokens.

\subsubsection{Geometry-Aware Reparameterization Module (GAR)}

\label{sec:gar}

With the global geometric constraints provided by GPI, GAR further performs fine-grained local alignment and feature reparameterization via cross-modal multi-head cross-attention. We first construct the key/value sequences:
\begin{equation}
\begin{aligned}
K_t &= F^{3D}_{t,\mathrm{raw}} W_K + b_K,\\
V_t &= F^{3D}_{t,\mathrm{raw}} W_V + b_V,
\end{aligned}
\label{eq:kv}
\end{equation}
and feed $Q^{\text{inj}}_t$ as the query sequence into the cross-attention layer.

\paragraph{Cross-modal multi-head cross-attention}
Let $H$ denote the number of heads and $d_h=d/H$ be the head dimension. For the $i$-th head, we have:
\begin{equation}
q_i = Q^{\text{inj}}_t W_i^q,\quad
k_i = K_t W_i^k,\quad
v_i = V_t W_i^v,
\label{eq:head_proj}
\end{equation}
where $W_i^q,W_i^k,W_i^v\in\mathbb{R}^{d\times d_h}$. The attention output is:
\begin{equation}
\mathrm{head}_i
=
\mathrm{softmax}\!\left(\frac{q_i k_i^\top}{\sqrt{d_h}}\right)v_i,
\label{eq:head_attn}
\end{equation}

We concatenate all heads and apply an output projection to obtain the aligned representations:
\begin{equation}
F^{\text{align}}_t
=
\mathrm{Concat}(\mathrm{head}_1,\dots,\mathrm{head}_H)W^O
\in\mathbb{R}^{N_{2D}\times d},
\label{eq:align}
\end{equation}

This operation allows each 2D token to adaptively retrieve and absorb the corresponding structural cues from the geometric tokens based on its semantic content and visual observations, resulting in a geometry-aware reparameterization in the feature space.

\paragraph{Gated residual fusion}
To preserve the semantic discrimination ability of CLIP while injecting geometric priors, we use a gated residual connection to fuse $F^{\text{align}}_t$ with the base query representations $Q^{\text{base}}_t$:
\begin{equation}
F^{\text{fuse}}_t
=
\alpha\cdot F^{\text{align}}_t
+
(1-\alpha)\cdot Q^{\text{base}}_t,
\qquad \alpha\in(0,1).
\label{eq:gated_fuse}
\end{equation}
where $\alpha$ is a learnable gating coefficient.
The resulting $F^{\text{fuse}}_t$ keeps the same token form as the original 2D tokens, and can be directly used as geometry-aligned 2D representations for the downstream navigation head.

\section{Experiments}
\label{sec:experiments}
\subsection{Experimental Setup}

We evaluate SpatialFly on the OpenUAV dataset. We report four standard UAV VLN metrics: Navigation Error (NE), Success Rate (SR), Oracle Success Rate (OSR), and Success weighted by Path Length (SPL)~\cite{feng2025vpnvisualpromptnavigation, xiang2025navr2dualrelationreasoninggeneralizable}. 

SpatialFly is implemented in PyTorch with Qwen2.5-3B as the language backbone, CLIP ViT-L/14 as the 2D visual encoder, and VGGT~\cite{wang2025vggtvisualgeometrygrounded} as the implicit geometry encoder. The original VGGT prediction heads are removed, and only its Transformer trunk tokens are used as geometric priors. The model predicts 3D waypoint increments and is optimized with a trajectory loss combining $L_1$ distance regression and cosine direction similarity. We train the model using LoRA fine-tuning with frozen encoders, DeepSpeed ZeRO-2, AdamW optimizer, and mixed precision on eight NVIDIA RTX 4090 GPUs.

\subsection{Results on the Test Seen Set}

Table~\ref{tab:seen_color_results} reports the results on the OpenUAV Test Seen set under both 25\% training data and 100\% training data settings. Under the low-data setting, SpatialFly consistently outperforms the strongest competing method across all difficulty splits. On the Full split, it reduces NE from 125.97\,m to 113.31\,m and improves SR from 14.39\% to 16.84\%. Similar gains are observed on the Easy and Hard splits, indicating that the injected geometric priors improve data efficiency and help the model learn more reliable spatial representations from limited training data.

Under the full-data setting, SpatialFly also achieves the best overall performance. Compared with LongFly, SpatialFly reduces NE from 60.02\,m to 58.05\,m and improves SR from 36.39\% to 38.54\% on the Full split. On the Hard split, SpatialFly further improves SR from 33.94\% to 37.33\% and SPL from 30.88\% to 33.22\%. Compared with the base model BS, SpatialFly brings larger improvements, especially on the Hard split, where NE decreases from 127.11\,m to 84.76\,m and SR increases from 15.91\% to 37.33\%. These results show that G2RA improves both general navigation accuracy and robustness in complex scenes.

\subsection{Results on the Test Unseen Set}

Table~\ref{tab:unseen_results} reports the generalization results on the Test Unseen set. SpatialFly consistently achieves better or competitive performance compared with previous methods. On the Full split, SpatialFly reduces NE from 91.84\,m to 87.82\,m and improves SR from 24.19\% to 25.46\% compared with LongFly. On the Easy split, it reduces NE from 69.16\,m to 64.49\,m and improves SR from 22.89\% to 25.17\%. On the Hard split, SpatialFly achieves the best SR, OSR, and SPL, reaching 25.71\%, 44.63\%, and 23.01\%, respectively.

Compared with the base model BS, SpatialFly shows more substantial gains on unseen environments. On the Full split, NE decreases from 106.08\,m to 87.82\,m and SR improves from 13.99\% to 25.46\%. On the Hard split, NE decreases from 133.49\,m to 108.56\,m and SPL improves from 10.52\% to 23.01\%. These results indicate that aligning 2D semantic tokens with implicit 3D geometric priors improves unseen-scene generalization and reduces spatial decision errors in challenging UAV VLN scenarios.

\begin{table}[t]
\caption{Results on the Test Seen Set across Full/Easy/Hard splits using NE↓ and SR/OSR/SPL(\%)↑.  
25\% (Low-Data) and 100\% (Full-Data) settings are visually separated for clarity.}
\label{tab:seen_color_results}
\centering
\small
\setlength{\tabcolsep}{2.5pt}
\renewcommand{\arraystretch}{1.12}

\begin{adjustbox}{max width=\linewidth}
\begin{tabular}{lcccccccccccc}
\toprule
\multirow{2}{*}{\textbf{Method}} &
\multicolumn{4}{c}{\textbf{Full}} &
\multicolumn{4}{c}{\textbf{Easy}} &
\multicolumn{4}{c}{\textbf{Hard}} \\
\cmidrule(lr){2-5} \cmidrule(lr){6-9} \cmidrule(lr){10-13}
& NE$\!\downarrow$ & SR$\!\uparrow$ & OSR$\!\uparrow$ & SPL$\!\uparrow$
& NE$\!\downarrow$ & SR$\!\uparrow$ & OSR$\!\uparrow$ & SPL$\!\uparrow$
& NE$\!\downarrow$ & SR$\!\uparrow$ & OSR$\!\uparrow$ & SPL$\!\uparrow$ \\
\midrule


\rowcolor{gray!20}
Human & \textbf{14.15} & \textbf{94.51} & \textbf{94.51} & \textbf{77.84}
      & \textbf{11.68} & \textbf{95.44} & \textbf{95.44} & \textbf{76.19}
      & \textbf{17.16} & \textbf{93.37} & \textbf{93.37} & \textbf{79.85} \\
\midrule
\rowcolor{blue!10}
\multicolumn{13}{l}{\textcolor{blue!60!black}{\textbf{25\% Training Data (Low-Data Setting)}}} \\
TravelUAV\cite{wang2024towards}        & 132.59 &11.59 &24.50 & 10.45 &85.75 &14.11 &30.35 &12.34 &181.19 &8.98 &18.43 &8.48 \\

OpenVLN\cite{lin2025openvlnopenworldaerialvisionlanguage} & 125.97 & 14.39 & 28.03 & 12.94
               & 87.96 & 15.22 & 30.64 & 13.31
               & 175.54 & 13.32 & 24.62 & 12.55 \\

\rowcolor{blue!5}
BS(Ours)

                &119.12	&13.12	&27.08	&9.50
&73.77	&15.96	&32.45	&11.21
&171.20	&9.85	&20.91	&7.55 \\

\rowcolor{blue!15}
SpatialFly (Ours) 
&\textbf{113.31}	&\textbf{16.84}	&\textbf{38.36}	&\textbf{14.58}
&\textbf{70.13}	&\textbf{18.60}	&\textbf{48.15}	&\textbf{16.12}
&\textbf{162.90}	&\textbf{14.82}	&\textbf{27.02}	&\textbf{12.81} \\

\midrule

\rowcolor{red!10}
\multicolumn{13}{l}{\textcolor{red!60!black}{\textbf{100\% Training Data (Full-Data Setting)}}} \\

Random Action   &222.20 &0.14 &0.21 &0.07 
                &142.07 &0.26 &0.39 &0.13 
                &320.12 &0.00 &0.00 &0.00 \\

Fixed Action    &188.61 &2.27 &8.16 &1.40 
                &121.36 &3.48 &11.48 &2.14 
                &270.69 &0.79 &4.09 &0.49 \\

CMA\cite{anderson2018vision} 
                &135.73 &8.37 &18.72 &7.90 
                &84.89 &11.48 &24.52 &10.68 
                &197.77 &4.57 &11.65 &4.51 \\

TravelUAV\cite{wang2024towards}
               & 106.28 & 16.10 & 44.26 & 14.30
               & 68.78  & 18.84 & 47.61 & 16.39
               & 152.04 & 12.76 & 40.16 & 11.76 \\

TravelUAV-DA\cite{wang2024towards}
               & 98.66  & 17.45 & 48.87 & 15.76
               & 66.40  & 20.26 & 51.23 & 18.10
               & 138.04 & 14.02 & 45.98 & 12.90 \\

NavFoM\cite{zhang2025embodiednavigationfoundationmodel}
               & 93.05 & 29.17 & 49.24 & 25.03
               & 58.98 & 32.91 & 53.16 & 27.87
               & 143.83 & 23.58 & 43.40 & 20.80 \\

\rowcolor{red!5}
LongFly\cite{jiang2025longflylonghorizonuavvisionandlanguage} 
        & 60.02 & 36.39 & 65.87 & 31.07 
        & 38.10 & 38.52 & 71.90 & 31.24 
        & 85.20 & 33.94 & 58.94 & 30.88 \\

\rowcolor{red!5}
BS(Ours)
               & 85.17 & 22.50 & 	44.64 &  19.12 & 48.65  &  28.23 & 	54.88 & 23.02 & 127.11 & 	15.91 & 32.88 & 14.64  \\

\rowcolor{red!15}
\textbf{SpatialFly (Ours)} 

&\textbf{58.05}	&\textbf{38.54}	&\textbf{68.62}	&\textbf{33.43}
&\textbf{34.41}	&\textbf{39.59}	&\textbf{74.14}	&\textbf{33.61}
&\textbf{84.76}	&\textbf{37.33}	&\textbf{62.27}	&\textbf{33.22} \\

\bottomrule

\end{tabular}
\end{adjustbox}
\end{table}

\begin{table}[t]
\caption{Results on the Test Unseen Set across Full/Easy/Hard splits.}
\label{tab:unseen_results}
\centering
\small
\setlength{\tabcolsep}{4pt}
\begin{adjustbox}{max width=\linewidth}
\begin{tabular}{lcccccccccccc}
\toprule
\multirow{2}{*}{Method} &
\multicolumn{4}{c}{\textbf{Full}} &
\multicolumn{4}{c}{\textbf{Easy}} &
\multicolumn{4}{c}{\textbf{Hard}} \\
\cmidrule(lr){2-5} \cmidrule(lr){6-9} \cmidrule(lr){10-13}
& NE$\!\downarrow$ & SR$\!\uparrow$ & OSR$\!\uparrow$ & SPL$\!\uparrow$
& NE$\!\downarrow$ & SR$\!\uparrow$ & OSR$\!\uparrow$ & SPL$\!\uparrow$
& NE$\!\downarrow$ & SR$\!\uparrow$ & OSR$\!\uparrow$ & SPL$\!\uparrow$ \\
\midrule
Random Action               & 225.64 & 0.06  & 0.06  & 0.06  & 164.66 & 0.19  & 0.19  & 0.19  & 280.58 & 0.00  & 0.00  & 0.00 \\
Fixed Action                & 193.30 & 1.76  & 5.36  & 1.09  & 140.33 & 3.19  & 8.08  & 1.88  & 245.96 & 0.85  & 3.08  & 0.55 \\
CMA\cite{anderson2018vision}                  & 147.27 & 4.98  & 12.41 & 4.74  & 102.54 & 8.03  & 17.52 & 7.52  & 191.30 & 2.76  & 7.53  & 2.71 \\
TravelUAV\cite{wang2024towards}              & 130.60 & 11.41 & 31.13 & 10.45 & 96.27  & 12.47 & 33.31 & 11.29 & 167.49 & 10.62 & 28.91 & 9.80 \\
NavFoM\cite{zhang2025embodiednavigationfoundationmodel}	& 118.34	& 15.63	& 30.46	& 14.21	& 89.77	& 16.98	& 32.22	& 15.35	& 155.69	& 14.35	& 27.79	& 13.16 \\

LongFly\cite{jiang2025longflylonghorizonuavvisionandlanguage} &91.84 &24.19 &43.86 &20.84 &69.16 &22.89 &43.24 &18.66 &112.02 &25.36 &44.41 &22.76\\

\midrule
\rowcolor{blue!10}
BS(Ours)     & 106.08 & 13.99 & 27.66 & 12.16 & 75.25  & 16.73 & 32.40 & 14.01 & 133.49 & 11.55 & 23.45 & 10.52 \\
\rowcolor{blue!20}
\textbf{SpatialFly(Ours)}       & \textbf{87.82}  & \textbf{25.46} & \textbf{44.41} & \textbf{21.76}
                     & \textbf{64.49}  & \textbf{25.17} & \textbf{44.18} & \textbf{20.36}
                     & \textbf{108.56} & \textbf{25.71} & \textbf{44.63} & \textbf{23.01} \\

\bottomrule
\end{tabular}
\end{adjustbox}
\end{table}

\subsection{Performance on the Test Unseen Map Set}

As shown in Table~\ref{tab:vln-results-um}, on the Test Unseen Map split, the proposed SpatialFly consistently outperforms the previous state-of-the-art method LongFly across all core metrics, further demonstrating its generalization capability under cross-scene structural distribution shifts.

On the Full subset, SpatialFly achieves 13.57\% SR, 30.38\% OSR, and 11.13\% SPL. Compared with LongFly (SR: 11.27\%, OSR: 30.27\%, SPL: 9.32\%), this corresponds to improvements of +2.30\%, +0.11\%, and +1.81\%, respectively. Meanwhile, NE is reduced from 108.32\,m to 104.05\,m (-4.27\,m). On the Easy subset, the advantage becomes more evident. SpatialFly achieves 16.06\% SR, improving by +3.10\% over LongFly, while SPL increases by +2.38\% and NE decreases by 5.62\,m. The OSR remains comparable to LongFly (34.31\%), suggesting that in relatively regular unseen environments, the proposed method enables more accurate goal convergence and improved path quality. On the more challenging Hard subset, SpatialFly also maintains superior performance, reaching 10.24\% SR, 25.12\% OSR, and 9.04\% SPL. Compared with LongFly, this corresponds to gains of +1.22\%, +0.24\%, and +1.06\%, respectively, while reducing NE by 2.47\,m. These results indicate that G2RA improves navigation success and path efficiency under unseen scene layouts. In structurally complex and long-horizon scenarios, explicitly modeling geometric consistency helps reduce cumulative spatial errors and enhances generalization across novel scene layouts.

\section{Ablation Studies}
\begin{table}[t]
\caption{Navigation performance on the Test Unseen Map split.}
\label{tab:vln-results-um}
\centering
\small
\setlength{\tabcolsep}{4pt}
\begin{adjustbox}{max width=\columnwidth}
\begin{tabular}{l*{12}{c}}
\toprule
\multirow{2}{*}{Method} &
\multicolumn{4}{c}{\textbf{Full}} &
\multicolumn{4}{c}{\textbf{Easy}} &
\multicolumn{4}{c}{\textbf{Hard}} \\
\cmidrule(lr){2-5}\cmidrule(lr){6-9}\cmidrule(lr){10-13}
& NE$\!\downarrow$ & SR$\!\uparrow$ & OSR$\!\uparrow$ & SPL$\!\uparrow$
& NE$\!\downarrow$ & SR$\!\uparrow$ & OSR$\!\uparrow$ & SPL$\!\uparrow$
& NE$\!\downarrow$ & SR$\!\uparrow$ & OSR$\!\uparrow$ & SPL$\!\uparrow$ \\
\midrule
Random Action   & 202.98 & 0.00 & 0.00 & 0.00 & 158.46 & 0.00 & 0.00 & 0.00 & 265.88 & 0.00 & 0.00 & 0.00 \\
Fixed Action     & 180.47 & 0.52 & 2.61 & 0.39 & 132.89 & 0.89 & 4.28 & 0.67 & 247.72 & 0.00 & 0.25 & 0.00 \\
CMA\cite{anderson2018vision}      & 141.68 & 2.30 & 10.02 & 2.16 & 102.29 & 3.57 & 14.26 & 3.33 & 197.35 & 0.50 & 4.03 & 0.50 \\
TravelUAV\cite{wang2024towards}  & {138.80} & {4.18} & {20.77} & {3.84}
         & {102.94} & {4.63} & {22.82} & {4.24}
         & {189.46} & {3.53} & {17.88} & {3.28} \\
NavFoM\cite{zhang2025embodiednavigationfoundationmodel}	& 125.10	& 6.30	& 18.95	& 5.68	& 102.41	& 6.77	& 20.07	& 6.04	& 170.58	& 5.36	& 15.71	& 4.97 \\
LongFly\cite{jiang2025longflylonghorizonuavvisionandlanguage} &108.32 &11.27 &30.27 &9.32 &78.56 &12.96 &34.31 &10.32 &148.10 &9.02 &24.88 &7.98 \\

\midrule
\rowcolor{blue!10}
BS(Ours) & 112.61 & 7.52 & 20.98 & 6.38 & 79.97 & 10.40 & 25.91 & 8.66 & 156.24 & 3.66 & 14.39 & 3.32 \\
\rowcolor{blue!20}
\textbf{SpatialFly(Ours)}                 & \textbf{104.05}  & \textbf{13.57} & \textbf{30.38} & \textbf{11.13}
                     & \textbf{72.94}  & \textbf{16.06} & \textbf{34.31} & \textbf{12.70}
                     & \textbf{145.63} & \textbf{10.24} & \textbf{25.12} & \textbf{9.04} \\

\bottomrule
\end{tabular}
\end{adjustbox}
\end{table}
\subsection{Ablation on Different Components}

\begin{table}[t]
\caption{Component ablation of SpatialFly on the Test Unseen split.}
\label{tab:ablation_2d3d}
\centering
\small
\setlength{\tabcolsep}{4pt}
\renewcommand{\arraystretch}{1.15}
\begin{adjustbox}{max width=\linewidth}
\begin{tabular}{lcccccccccccccccc}
\toprule
\multirow{2}{*}{Variant}
& \multirow{2}{*}{\centering 2D Sem.}
& \multirow{2}{*}{\centering 3D Geo.}
& \multirow{2}{*}{\centering Fusion}
& \multirow{2}{*}{\centering Geo Inject}
& \multicolumn{4}{c}{\textbf{Full}}
& \multicolumn{4}{c}{\textbf{Easy}}
& \multicolumn{4}{c}{\textbf{Hard}} \\
\cmidrule(lr){6-9} \cmidrule(lr){10-13} \cmidrule(lr){14-17}
& & & & 
& NE$\downarrow$ & SR$\uparrow$ & OSR$\uparrow$ & SPL$\uparrow$
& NE$\downarrow$ & SR$\uparrow$ & OSR$\uparrow$ & SPL$\uparrow$
& NE$\downarrow$ & SR$\uparrow$ & OSR$\uparrow$ & SPL$\uparrow$ \\
\midrule

2D-only
& $\checkmark$ & $\times$ & -- & $\times$
& 105.96 & 19.36 & 38.56 & 16.52
& 78.15  & 19.14 & 40.03 & 15.49
& 130.93 & 19.54 & 37.23 & 17.43 \\

3D-only
& $\times$ & $\checkmark$ & -- & $\times$
& 112.15 & 16.89 & 37.18 & 14.80
& 78.68  & 14.99 & 38.42 & 12.45
& 141.91 & 18.57 & 36.07 & 16.88 \\

Concat Fusion
& $\checkmark$ & $\checkmark$ & Concat & $\times$
& 115.93 & 16.76 & 34.15 & 13.84
& 81.53  & 15.40 & 35.21 & 11.93
& 146.51 & 17.98 & 33.22 & 15.54 \\

No Geo Inject
& $\checkmark$ & $\checkmark$ & G$^{2}$RA & $\times$
& 95.66  & 23.44 & 42.34 & 19.43
& 69.29  & 23.03 & 43.11 & 18.08
& 119.11 & 23.81 & 41.67 & 20.64 \\

\rowcolor{blue!10}
\textbf{SpatialFly (Full)}
& $\checkmark$ & $\checkmark$ & G$^{2}$RA & $\checkmark$
& \textbf{87.82}  & \textbf{25.46} & \textbf{44.41} & \textbf{21.76}
& \textbf{64.49}  & \textbf{25.17} & \textbf{44.18} & \textbf{20.36}
& \textbf{108.56} & \textbf{25.71} & \textbf{44.63} & \textbf{23.01} \\
\bottomrule
\end{tabular}
\end{adjustbox}
\end{table}

To better understand the contribution of each component in SpatialFly, we conduct component-level ablation experiments on the Test Unseen split, as shown in Table~\ref{tab:ablation_2d3d}. We analyze both the necessity of the semantic/geometric branches and the effectiveness of the key design choices in the proposed fusion mechanism.

\subsubsection{Effect of Different Branches} 

We first evaluate whether both the semantic and geometric branches are necessary for robust UAV VLN. The results show that using only the 2D semantic branch (2D-only) or only the 3D geometric branch (3D-only) leads to clear performance degradation. On the Full split, the SR drops to 19.36\% for 2D-only and 16.89\% for 3D-only, compared with 25.46\% achieved by the full model, while NE increases noticeably in both cases. These results suggest that a single modality is insufficient to jointly capture semantic understanding and spatial structure, and that both branches are necessary for complex navigation.

\subsubsection{Effect of Key Design Choices} 

We further evaluate several key design choices in the proposed fusion module. Replacing the proposed G2RA operator with simple feature concatenation (Concat Fusion) results in substantial performance degradation. On the Full split, SR and SPL decrease to 16.76\% and 13.84\%, respectively, while NE increases to 115.93\,m. This indicates that naive concatenation is insufficient to model the relationship between 2D semantic and 3D geometric representations. Moreover, when keeping the G2RA fusion structure but removing geometric prior injection (No GPI), consistent performance degradation is still observed. On the Hard split, SR decreases from 25.71\% to 23.81\%, and NE increases from 108.56\,m to 119.11\,m. This suggests that GPI is particularly beneficial in complex scenarios.

\begin{figure}[t]
    \centering
    \includegraphics[width=0.9\linewidth]{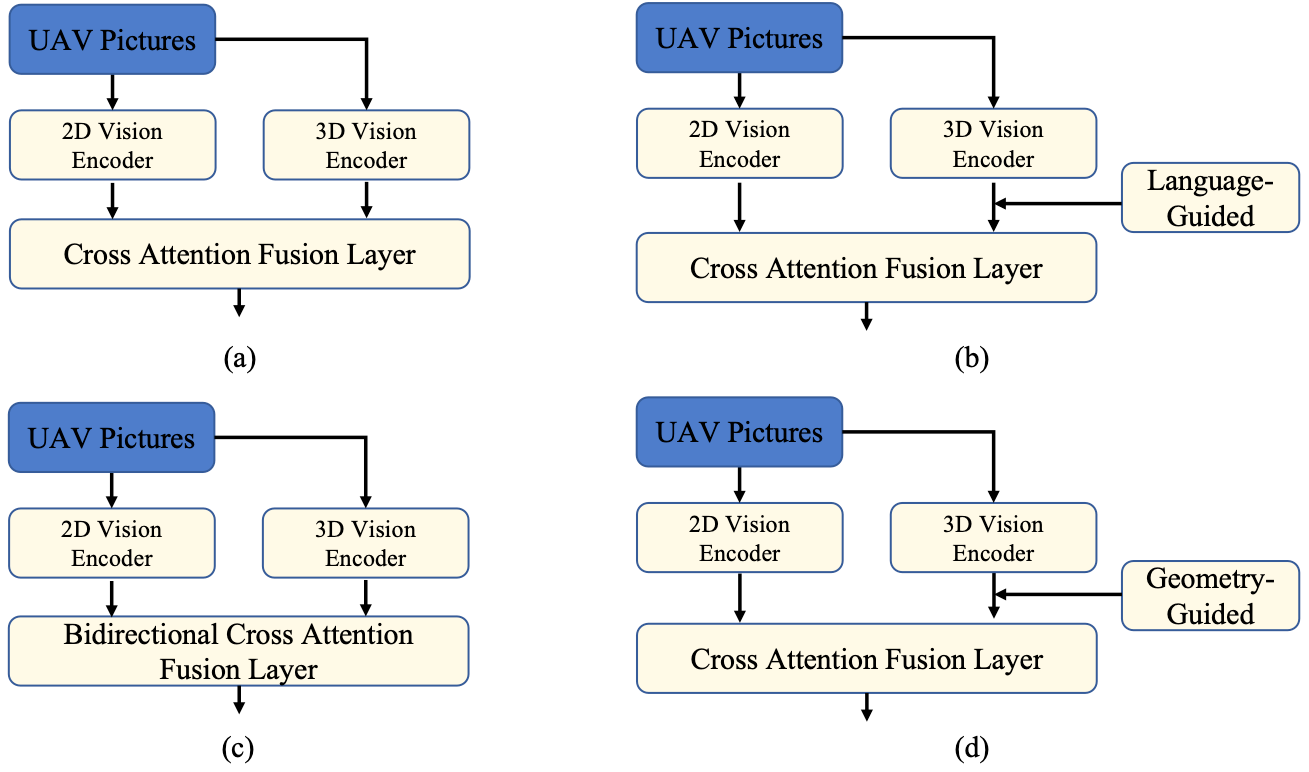}

    \caption{Fusion architecture variants implemented for comparison.
(a) Unidirectional cross-attention baseline following Evo-0~\cite{lin2025evo0visionlanguageactionmodelimplicit}.
(b) Language-modulated cross-attention inspired by CLIPORT~\cite{shridhar2021cliportpathwaysroboticmanipulation}.
(c) Bidirectional interaction mechanism following~\cite{nguyen-tran-etal-2022-bi}.
(d) The proposed SpatialFly with G2RA.}
    \label{fig:cross_attention_diff}
\end{figure}

\subsection{Comparison of Fusion Architecture Designs}

\begin{table}[t]
\caption{Comparison with different fusion architectures on the Test Unseen split.}
\label{tab:fusion_compare_unseen}
\centering
\small
\setlength{\tabcolsep}{4pt}
\begin{adjustbox}{max width=\columnwidth}
\begin{tabular}{lcccccccccccc}
\toprule
\multirow{2}{*}{Method} &
\multicolumn{4}{c}{\textbf{Full}} &
\multicolumn{4}{c}{\textbf{Easy}} &
\multicolumn{4}{c}{\textbf{Hard}} \\
\cmidrule(lr){2-5}\cmidrule(lr){6-9}\cmidrule(lr){10-13}
& NE$\!\downarrow$ & SR$\!\uparrow$ & OSR$\!\uparrow$ & SPL$\!\uparrow$
& NE$\!\downarrow$ & SR$\!\uparrow$ & OSR$\!\uparrow$ & SPL$\!\uparrow$
& NE$\!\downarrow$ & SR$\!\uparrow$ & OSR$\!\uparrow$ & SPL$\!\uparrow$ \\
\midrule

Cross-Attention(Evo-0-style)~\cite{lin2025evo0visionlanguageactionmodelimplicit}
& 100.13 & 22.68 & 41.34 & 18.73
&  71.15 & 21.55 & 40.56 & 16.43
& 125.89 & 23.69 & 42.02 & 20.76 \\

Language-Guided(CLIPORT-style)~\cite{shridhar2021cliportpathwaysroboticmanipulation}
&  92.10 & 23.57 & 42.41 & 20.26
&  69.02 & 22.22 & 40.96 & 18.04
& 112.64 & 24.76 & 43.69 & 22.23 \\

Bidirectional Cross-Attention~\cite{nguyen-tran-etal-2022-bi}
&  91.63 & 23.00 & 39.26 & 19.14
&  67.32 & 21.21 & 39.87 & 16.48
& 113.14 & 24.58 & 38.72 & 21.49 \\

\rowcolor{blue!10}
\textbf{SpatialFly (Ours)}
& \textbf{87.82}  & \textbf{25.46} & \textbf{44.41} & \textbf{21.76}
& \textbf{64.49}  & \textbf{25.17} & \textbf{44.18} & \textbf{20.36}
& \textbf{108.56} & \textbf{25.71} & \textbf{44.63} & \textbf{23.01} \\
\bottomrule
\end{tabular}
\end{adjustbox}
\end{table}

To evaluate the rationality and effectiveness of the G2RA design, we compare SpatialFly with several representative fusion architectures inspired by prior work, as illustrated in Fig.~\ref{fig:cross_attention_diff}(a)-(d). The results are reported in Table~\ref{tab:fusion_compare_unseen}. These comparison models are structural reference implementations and are used only to analyze the effect of different fusion strategies.

Structure (a) follows the unidirectional cross-attention mechanism adopted in Evo-0~\cite{lin2025evo0visionlanguageactionmodelimplicit} and serves as a standard cross-modal attention baseline. Structure (b) draws on the language-modulated cross-attention design in CLIPORT~\cite{shridhar2021cliportpathwaysroboticmanipulation}, where language features are used to regulate the cross-modal interaction process. Structure (c) follows a bidirectional interaction mechanism~\cite{nguyen-tran-etal-2022-bi}, enabling two-way attention flow between 2D and 3D features. Although these variants improve feature interaction to different extents, their gains remain limited, especially under more challenging settings. In contrast, the proposed SpatialFly explicitly introduces geometry-guided constraints into the fusion process and achieves the best overall performance. On the Full split, it reaches 25.46\% SR, 44.41\% OSR, and 21.76\% SPL, with NE reduced to 87.82\,m. Overall, these results suggest that attention-based interaction or semantic modulation alone is insufficient, while explicit geometric constraints help reduce error accumulation and improve path quality.

\subsection{Trajectory Alignment and Smoothness Analysis}

Beyond the conventional navigation metrics~\cite{zhang2026apexdecoupledmemorybasedexplorer}, we further analyze trajectory quality from the perspectives of path alignment and motion smoothness. Specifically, NDTW measures the global similarity between the predicted trajectory and the reference route, while SDTW further incorporates task success by assigning zero to failed trajectories~\cite{wang2026cityseeker}. To evaluate motion quality, we compute the mean and variance of turning angles between consecutive trajectory segments, where lower values indicate smoother trajectory transitions and more stable flight behavior.

As shown in Table~\ref{tab:traj_alignment_smoothness}, SpatialFly consistently outperforms LongFly in both NDTW and SDTW across the Full, Easy, and Hard splits, indicating that its predicted trajectories are more faithfully aligned with the global shape of the reference paths. The gains are particularly notable on the Hard split, where NDTW and SDTW improve by 2.57 and 1.10 points, respectively, suggesting that the proposed spatial guidance is especially beneficial in complex scenes that require stronger route-level spatial reasoning. Meanwhile, SpatialFly achieves lower smoothness mean and variance on all three splits, demonstrating reduced turning fluctuations and more stable motion generation. These results show that the G2RA improves not only navigation success, but also the path alignment and motion quality of the predicted trajectories in continuous UAV navigation.

\begin{table}[t]
\caption{Comparison of trajectory alignment and smoothness metrics on the Test Unseen split.}
\label{tab:traj_alignment_smoothness}
\centering
\small
\setlength{\tabcolsep}{4pt}
\begin{adjustbox}{max width=\columnwidth}
\begin{tabular}{lcccccccccccc}
\toprule
\multirow{2}{*}{Method} &
\multicolumn{4}{c}{\textbf{Full}} &
\multicolumn{4}{c}{\textbf{Easy}} &
\multicolumn{4}{c}{\textbf{Hard}} \\
\cmidrule(lr){2-5}\cmidrule(lr){6-9}\cmidrule(lr){10-13}
& Var$\!\downarrow$ & NDTW$\!\uparrow$ & SDTW$\!\uparrow$ & Mean$\!\downarrow$
& Var$\!\downarrow$ & NDTW$\!\uparrow$ & SDTW$\!\uparrow$ & Mean$\!\downarrow$
& Var$\!\downarrow$ & NDTW$\!\uparrow$ & SDTW$\!\uparrow$ & Mean$\!\downarrow$ \\
\midrule

LongFly\cite{jiang2025longflylonghorizonuavvisionandlanguage}
& 129.39 & 9.31 & 5.57 & 8.34
& 133.29 & 9.09 & 5.31 & 8.62
& 118.11 & 9.51 & 5.81 & 8.25 \\

\rowcolor{blue!10}
\textbf{SpatialFly (Ours)}
& \textbf{108.11} & \textbf{11.20} & \textbf{6.50} & \textbf{7.37}
& \textbf{114.10} & \textbf{10.20} & \textbf{6.04} & \textbf{7.49}
& \textbf{105.21} & \textbf{12.08} & \textbf{6.91} & \textbf{7.32} \\

\bottomrule
\end{tabular}
\end{adjustbox}
\end{table}

\begin{table}[t]
\centering
\caption{Sensitivity analysis of G2RA hyperparameters on the Test Unseen split.
$\eta$ denotes the geometric prior injection strength and $\alpha$ denotes the cross-modal fusion weight.}
\label{tab:eta_alpha_sensitivity}

\resizebox{\columnwidth}{!}{%
\begin{tabular}{c c c|cccc|cccc}
\toprule
\multirow{2}{*}{\textbf{ID}} &
\multirow{2}{*}{$\boldsymbol{\eta}$} &
\multirow{2}{*}{$\boldsymbol{\alpha}$} &
\multicolumn{4}{c|}{\textbf{Full}} &
\multicolumn{4}{c}{\textbf{Easy}} \\
\cmidrule(lr){4-7} \cmidrule(lr){8-11}
 & & &
NE$\!\downarrow$ & SR$\!\uparrow$ & OSR$\!\uparrow$ & SPL$\!\uparrow$ &
NE$\!\downarrow$ & SR$\!\uparrow$ & OSR$\!\uparrow$ & SPL$\!\uparrow$ \\
\midrule
1 & 0.0 & 0.2 &120.47	&12.16	&25.08	&11.36	&84.30	&13.39	&29.85	&11.99 \\
2 & 0.0 & 0.5 &89.29	&23.34	&41.74	&22.07	&66.98	&25.03	&43.78	&19.81 \\
3 & 0.0 & 0.8 &160.28	&5.36	&12.04	&5.15	&95.42	&8.43	&20.75	&8.01 \\
\midrule
4 & 0.5 & 0.2 &88.56	&24.89	&43.56	&21.22	&65.06	&23.02	&43.52	&18.68 \\
5 & 0.5 & 0.5 &87.82	&25.46	&44.41	&21.76	&64.49	&25.17	&44.18	&20.36 \\
6 & 0.5 & 0.8 &155.73	&5.48	&12.09	&5.26	&94.43	&8.17	&20.21	&7.70 \\
\midrule
7 & 1.0 & 0.2 &104.42	&17.77	&37.11	&15.28	&76.62	&16.47	&37.62	&13.39 \\
8 & 1.0 & 0.5 &93.13	&20.35	&40.83	&17.64	&70.70	&20.88	&41.77	&17.25 \\
9 & 1.0 & 0.8 &93.61	&22.05	&42.72	&19.04	&68.59	&20.22	&41.64	&16.58 \\

\midrule
\rowcolor{blue!10}
\textbf{Best} & $\eta^\star = 0.5$ & $\alpha^\star = 0.5$
& \textbf{87.82}  & \textbf{25.46} & \textbf{44.41} & \textbf{21.76}
                     & \textbf{64.49}  & \textbf{25.17} & \textbf{44.18} & \textbf{20.36} \\
\bottomrule
\end{tabular}%
}

\end{table}

\subsection{Sensitivity Analysis on $\eta$ and $\alpha$}

To analyze the influence of the geometric prior injection strength $\eta$ and the cross-modal fusion weight $\alpha$, we conduct systematic experiments with different $(\eta,\alpha)$ configurations on the Full and Easy splits. The results are reported in Table~\ref{tab:eta_alpha_sensitivity}.

As shown in Table~\ref{tab:eta_alpha_sensitivity}, the model shows a clear sensitivity to both $\eta$ and $\alpha$. The best overall performance is obtained when $\eta=0.5$ and $\alpha=0.5$, achieving 87.82,m NE, 25.46\% SR, 44.41\% OSR, and 21.76\% SPL on the Full split. The same setting also performs best on the Easy split, with 64.49,m NE, 25.17\% SR, and 20.36\% SPL, indicating that moderate geometric injection and balanced fusion are beneficial across different scene difficulties.

When $\eta=0$, the performance drops, suggesting that removing geometric guidance weakens spatial alignment. In contrast, overly strong injection also harms performance. For example, when $\eta=1.0$ and $\alpha=0.2$, SR decreases to 17.77\% and NE increases to 104.42,m on the Full split. Similarly, a large fusion weight such as $\alpha=0.8$ leads to obvious degradation, indicating that excessive cross-modal fusion may disturb the balance between semantic and geometric representations.

Overall, these results demonstrate that $\eta$ and $\alpha$ have a coupled effect on navigation performance. G2RA achieves stable and controllable behavior under moderate geometric injection and balanced fusion, while insufficient guidance or excessive fusion can degrade performance.

\begin{figure}[t]
    \centering
    \includegraphics[width=1\linewidth]{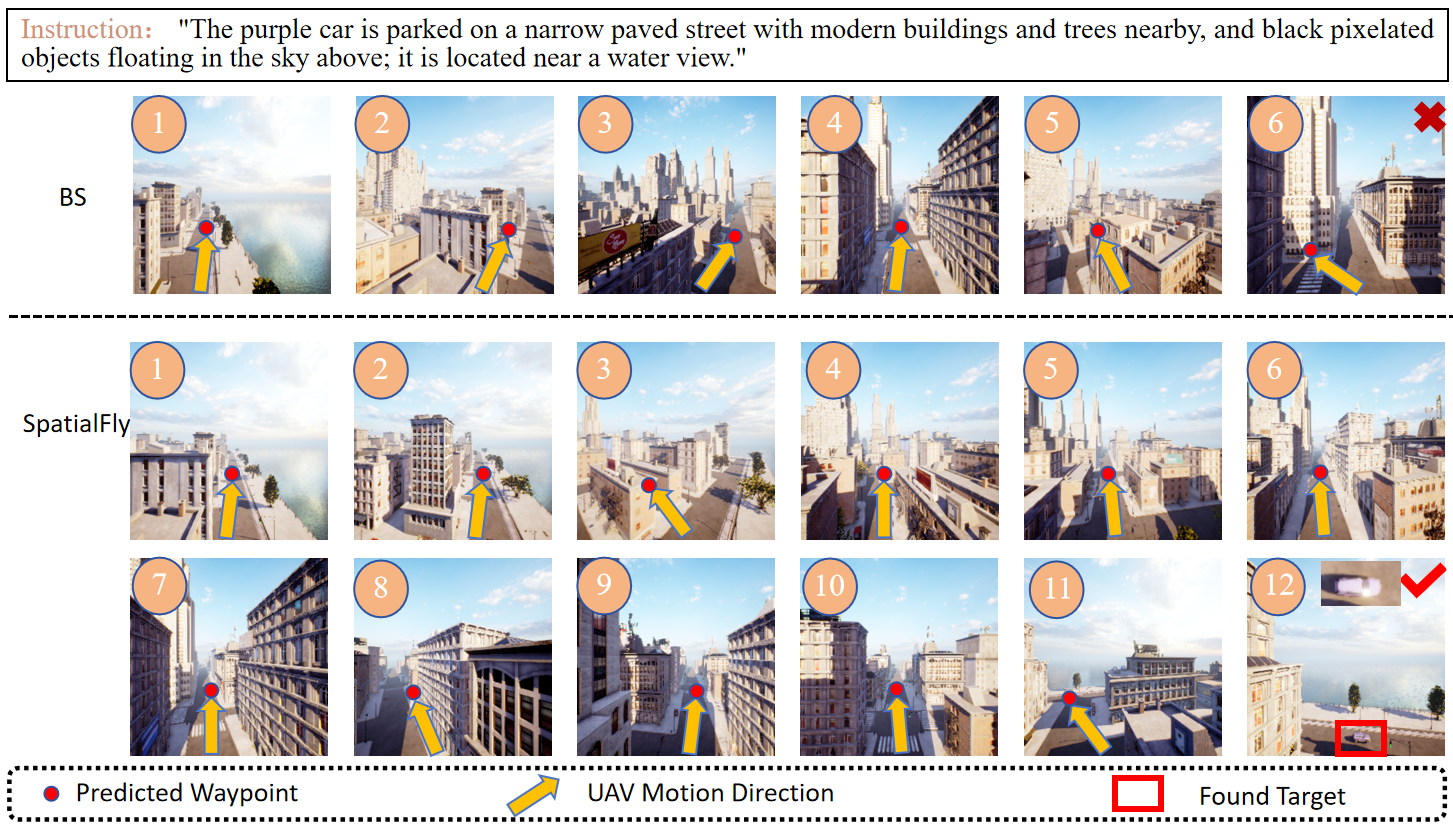}
    \caption{First-person visual comparison of UAV navigation.
Red dots indicate predicted waypoints, yellow arrows denote UAV motion directions, and red boxes mark the target. SpatialFly shows more accurate and stable target-oriented navigation than BS.}
    \label{fig:visual-1}
\end{figure}

\begin{figure}[t]
    \centering
    \includegraphics[width=1\linewidth]{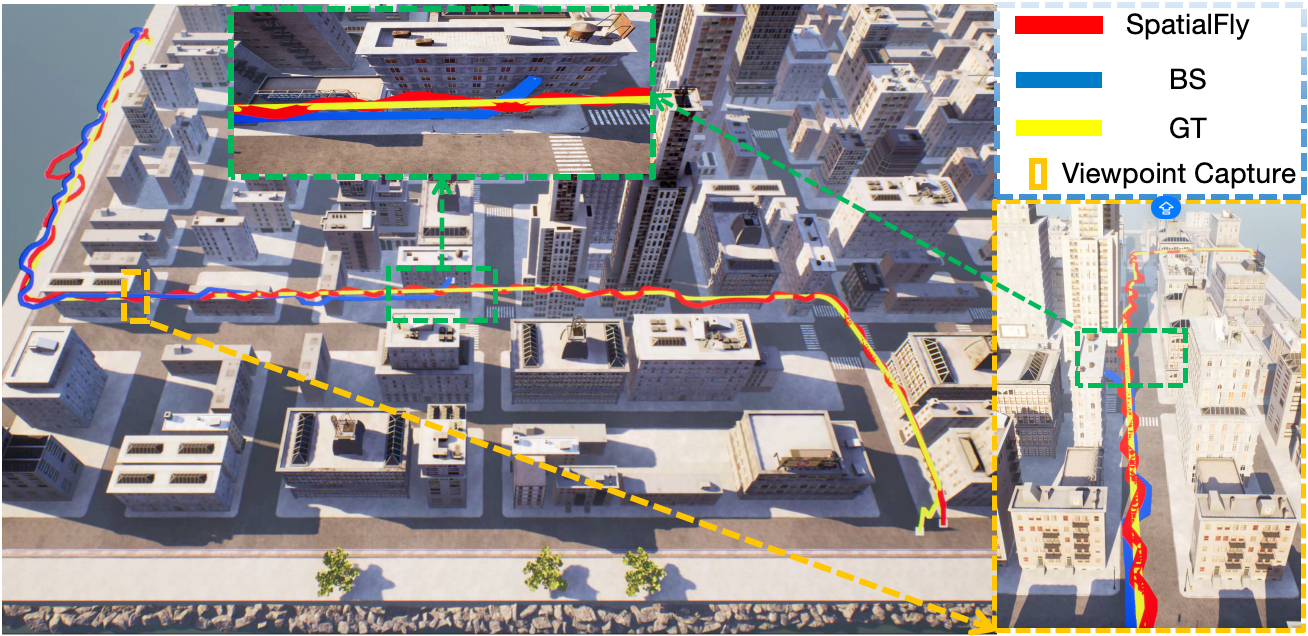}
    \caption{Third-person trajectory visualization of UAV navigation.
Red, blue, and yellow curves denote the trajectories of SpatialFly, BS, and ground truth, respectively. SpatialFly follows a trajectory closer to the ground truth than BS.}
    \label{fig:pic-fin-visual}
\end{figure}

\begin{figure}[t]
    \centering
    \includegraphics[width=\linewidth]{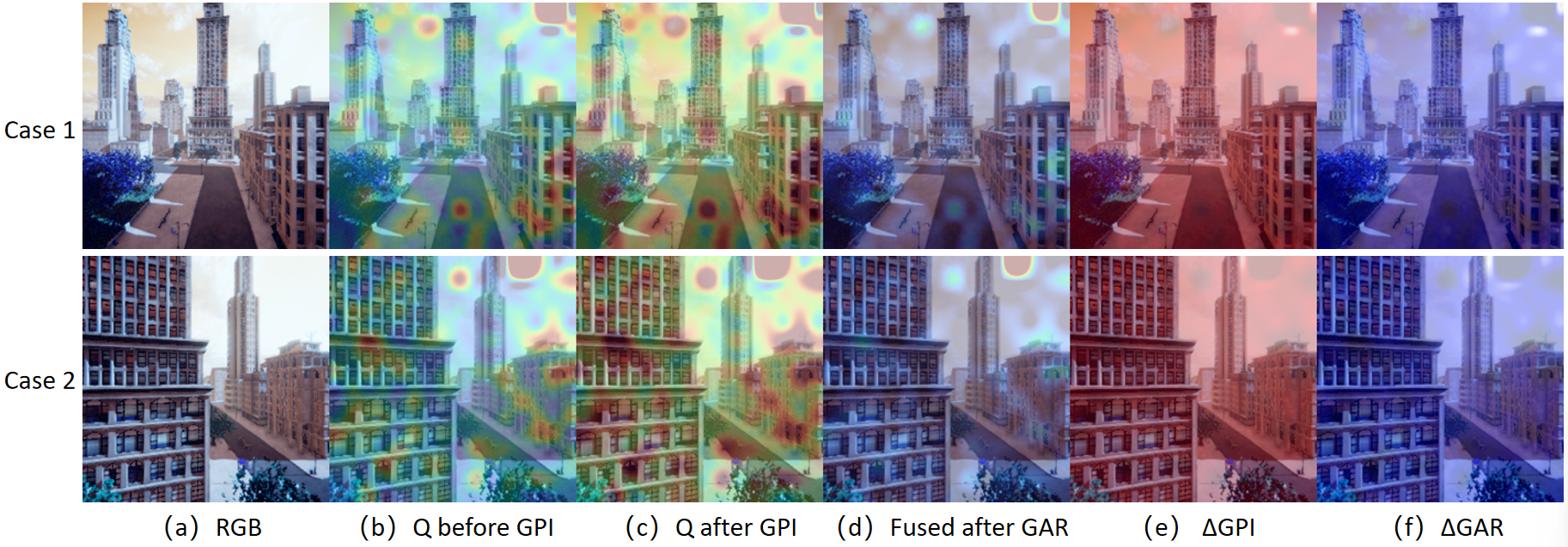}
    \caption{Visualization of image-space token responses before and after G2RA in two representative urban scenes. 
    (a) Input RGB image.
    (b) Original 2D token response before geometric prior injection (Q before GPI), which can be regarded as the original 2D representation without G2RA. 
    (c) 2D token response after geometric prior injection (Q after GPI). 
    (d) Fused response after geometry-aware reparameterization (Fused after GAR). 
    (e) Response change introduced by GPI ($\Delta$GPI = Q after GPI $-$ Q before GPI). 
    (f) Response change introduced by GAR ($\Delta$GAR = Fused after GAR $-$ Q after GPI).  
    Warmer colors in the absolute maps indicate stronger feature responses, while red/blue in the difference maps denote enhanced/suppressed responses relative to the previous stage.}
    \label{fig:geom-vis}
\end{figure}

\subsection{Qualitative Visualization and Analysis} 

\subsubsection{Navigation Behavior Visualization}

As shown in Fig.~\ref{fig:visual-1} and Fig.~\ref{fig:pic-fin-visual}, we provide a qualitative comparison between SpatialFly and the baseline BS on a representative language-guided target search task. The instruction requires the UAV to locate a target vehicle parked on a narrow street in a waterfront urban environment.

From the first-person perspective (Fig.~\ref{fig:visual-1}), BS can recognize some salient semantic cues in the scene, such as the waterfront area, but its subsequent viewpoint progression and waypoint prediction are not sufficiently stable. As a result, it fails to continuously focus on the target-related region and eventually misses the goal. In contrast, SpatialFly shows more stable target-oriented behavior. It gradually adjusts its search toward the street area between buildings and finally succeeds in locating the target.

The third-person trajectory visualization (Fig.~\ref{fig:pic-fin-visual}) further shows that the trajectory generated by SpatialFly is closer to the ground-truth (GT) path, with better continuity and smoothness. By comparison, the BS trajectory exhibits larger deviations from the GT path, together with unnecessary local drifts. This suggests that SpatialFly can better utilize the spatial structure of the scene to maintain more reasonable navigation progress, rather than relying mainly on local visual cues. These observations indicate that SpatialFly yields more reliable target-oriented navigation. We next visualize the image-space responses of 2D tokens before and after G2RA.

\subsubsection{Geometry-Guided Response Visualization} 

To further understand why SpatialFly yields more reliable navigation behavior, we visualize the image-space responses of 2D tokens before and after G2RA, as shown in Fig.~\ref{fig:geom-vis}. These visualizations should be interpreted as image-space token response maps rather than feature-space clustering results. Here, the response of Q before GPI can be regarded as the original 2D visual representation without G2RA. After GPI, the responses become more structured around scene layouts such as building facades, roadway regions, and structural transition areas, suggesting that the injected geometric priors reorganize the spatial distribution of 2D tokens. The $\Delta$GPI maps further show that these changes mainly occur around structure-related regions rather than being uniformly distributed across the image. 

After GAR, many localized activations are further suppressed, leading to a smoother and more regularized fused representation, as also reflected by the predominantly negative regions in the $\Delta$GAR maps. These observations indicate that GPI introduces geometric guidance into the original 2D representation, while GAR further filters redundant responses and regularizes the fused features. 

This leads to a more stable spatial representation for downstream navigation and provides qualitative support for alleviating the structural mismatch between 2D visual perception and 3D trajectory decision-making.

\section{Conclusion}
\label{sec:conclusion}

This paper proposes SpatialFly, a geometry-guided spatial representation framework for
UAV VLN in complex 3D environments. To mitigate the mismatch between 2D perception and 3D decision spaces, SpatialFly injects implicit 3D geometric priors into visual tokens and reparameterizes visual tokens through geometry-conditioned cross-modal attention and gated residual fusion. Experiments on Test Seen, Test Unseen, and Test Unseen Map show consistent improvements. Compared with unidirectional, bidirectional, and language-modulated attention variants, G2RA is more robust in structurally complex cases. Ablation and sensitivity studies further support the effectiveness and stability of the proposed design. In addition, trajectory-level analyses show that SpatialFly produces paths that are better aligned with reference trajectories while maintaining smoother and more stable motion behavior. Overall, explicitly modeling geometric consistency improves generalization and robustness in unseen environments.

\bibliographystyle{IEEEtran}
\bibliography{references}
\end{document}